\documentclass[twocolumn,10pt,abstract=on]{wmrDoc}

\usepackage{epsfig} %% for loading postscript figures
\usepackage{hyperref}
\hypersetup{colorlinks,linkcolor={black},citecolor={black},urlcolor={black}}  
\usepackage{graphicx} % Required for including images
\usepackage[font=small,labelfont=bf]{caption}
\hypersetup{colorlinks=true,urlcolor=black}

\title{Offensive Language and Hate Speech Detection with Deep Learning and Transfer Learning}

\author{
    Bencheng Wei\\
    20bw3@queensu.ca
    \\\\
    Jason Li\\
    20dl12@queensu.ca
    \and
    Ajay Gupta\\
    20ag22@queensu.ca
    \\\\
    Hafiza Umair\\
    20hku@queensu.ca
    \and
    Atsu Vovor\\
    19av27@queensu.ca
    \\\\
    Natalie Durzynski\\
    20nd5@queensu.ca
}

\begin{document}

\maketitle    
\begin{abstract}
{\it \textbf{Abstract:} Toxic online speech has become a crucial problem nowadays due to an exponential increase in the use of internet by people from different cultures and educational backgrounds. Differentiating if a text message belongs to hate speech and offensive language is a key challenge in automatic detection of toxic text content. In this paper, we propose an approach to automatically classify tweets into three classes: Hate, offensive and Neither. Using public tweet data set, we first perform experiments to build BI-LSTM models from empty embedding and then we also try the same neural network architecture with pre-trained Glove embedding. Next, we introduce a transfer learning approach for hate speech detection using an existing pre-trained language model BERT (Bidirectional Encoder Representations from Transformers), DistilBert (Distilled version of BERT) and GPT-2 (Generative Pre-Training). We perform hyper parameters tuning analysis of our best model (BI-LSTM) considering different neural network architectures, learn-ratings and normalization methods etc. After tuning the model and with the best combination of parameters, we achieve over 92 percent accuracy upon evaluating it on test data. We also create a class module which contains main functionality including text classification, sentiment checking and text data augmentation. This model could serve as an intermediate module between user and Twitter.}
\end{abstract}

\section{Introduction}

With most communication and advertising around brands going digital, companies have become more concerned about hate speech content than ever before. Hate speech content online can be defined as written messages with pejorative or discriminatory language. Although brands have control over the content they release and post online through their website and social media channels, they do not have full control over what online users post or comment surrounding their brand.

We have seen the evolution of what constitutes brand marketing over the past decade – the emergence of ‘Community Manager’ and ‘Social Content Manager’ type roles reflect the need for monitoring the communication around brand’s online content. This has traditionally focused on community engagement meaning driving responses or posts around brand content online. This in turn gave rise to social monitoring platforms like Olapic, Meltwater, or Stackla, that focus on driving engagement and brand awareness online. However, with the rise of hate speech, the role of brand marketing continues to take on responsibility; today’s Community Managers, and other similarly focused roles, must also monitor their brand’s digital channels to identify any potential hate speech.

When done without any tool in place, hate speech or offensive language detection is a manually intensive process that requires a lot of time and dedicated resources. A Community Manager would not have the bandwidth necessary to thoroughly track all brand associated content to detect any hate speech. They may in turn need to add additional resources on their team to address the growing need for this type of digital content control. It is also important to note that the human’s ability to accurately detect hate speech content is dependent on a multitude of factors including their energy levels, reading abilities, and personal biases of what constitutes negative content. The use of AI will allow for automation of this manually intensive process to create efficiencies and boost accuracy of the detection. We proposes an algorithm to automatically detect hate speech from Twitter content to help brands control their brand image online, cut down on time spent manually monitoring online, and mitigate any risks associated of being linked to hate speech. 

% \includegraphics[width=0.28\textwidth]{figure/Architettura1.png}
% \captionof{figure}{DL architecture sequence labeling}

\section{Data Summary and Preprocessing}

In this section, we will briefly summarize the data source and known tasks of the NLP will be dealt with including Sentiment analysis, Text cleansing, Text Augmentation, etc.

\subsection{Data Source}
In this paper, we are using public tweet data.

The data is stored as a CSV and comprised of 24,783 tweets. These tweets are labelled as Hate Speech, Offensive Language, or Neither by CrowdFlower (CF) users. This data is represented in tabular form with 24,783 rows and 6 columns; count, hate-speech, offensive-language, neither, class, and tweet. 

\subsection{Data Trends and Imbalanced Data}
The tweets were classified by groups of 3 to 9 CF users, however, a vast majority (92\%) of the tweets were judged buy a group of 3 CF users. The tweets were given the class which received the highest number of votes from the CF users:
    \begin{description}
    \item [$\bullet$] 6\% of the tweets were hate-speech – Class 0
    \item [$\bullet$] 77\% of the tweets were offensive-language – Class 1
    \item [$\bullet$] 17\% of the tweets were neither – Class 2
    \end{description}

More than 2/3 of the tweets were classified as class 1. This implies that the data is imbalanced. Furthermore, we noticed that the words used frequently in class 0 and class 1 are similar, as shown below, which makes it even more challenging to train the model to accurately classify tweets as class 0 or class 1. The most commonly used words in class 2 are clearly different from the words used in class 0 and class 1. 

\subsection{Sentiment Analysis}

Sentiment analysis is the process of detecting positive or negative sentiment in text. It is often used by businesses to detect sentiment in social data, gage brand reputation, and understand customers. Sentiment analysis is extremely important because it helps businesses to quickly understand the overall opinions of their customers. By automatically sorting the sentiment behind reviews, social media conversations, and more, you can make faster and more accurate decisions.

A social media sentiment analysis tells you how people feel about your brand online. Rather than a simple count of mentions or comments, sentiment analysis considers emotions and opinions. It involves collecting and analyzing information in the posts people share about your brand on social media.

Here we use the sentiment function of textblob to analyze the sentiment of tweet data. Textblob returns two properties, polarity, and subjectivity when we run it on the text data. Polarity is float which lies in the range of [-1,1] where 1 represents positive statement and -1 shows a negative statement. Subjective sentences generally refer to personal opinion, emotion or judgment whereas objective sentences refer to factual information. Subjectivity is also a float which lies in the range of [0,1].

Here we apply textblob to analyze the tweet with simple preprocessing. By doing this, we can see there is no significance different between different classes for subjectivity. Hate and offensive classes have about 0.4 subjective score compares to Neither which is about 0.3. In terms of polarity, we do see obvious difference between classes. Hate class has negative polarity compared to the Neither class which have about 0.08 positive polarity. Offensive class is neutral at this case with around 0 polarity. (Please Refer to Figure 1)

In Summary, Hate class usually has more negative sentiment (Polarity) compared to Neither and Offensive. This could prove to be important information that helps us to differenitate Hate class from Offensive and Neither class.

\begin{figure}
\centering
  \centering
  \includegraphics[width=0.4\linewidth]{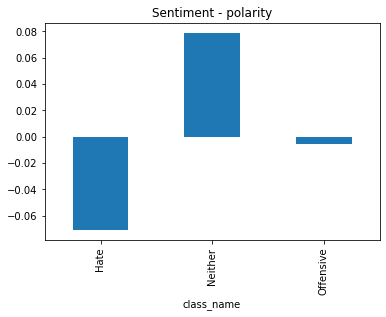}
  \label{fig:sub1}
  \centering
  \includegraphics[width=0.4\linewidth]{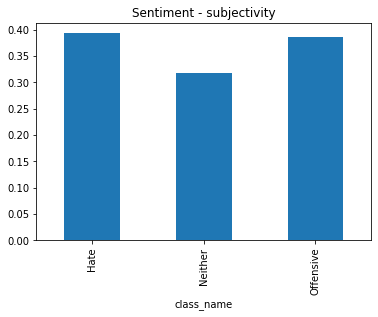}
  \label{fig:1}
\caption{Sentiment Analysis}
\label{fig:test}
\end{figure}

\subsection{Data Cleansing: Missing data and Contradictions Check}
Before we go and preprocess our tweet data and fit the data into model, we want to make sure our dataset is “clean”. We must ensure that there is no missing data, no duplication and no contradictions in the tweet data. 

After our initial analysis, there are no missing values in any of the columns. We compared unique values in the data to the total values in each row to check for duplicates. The table below shows that all the values in the ‘tweet’ column are unique, which indicates that there are no duplicate values in the data.

In term of contradictions, we want to make sure there is no equal vote on the classes, no label contradiction, etc. The tweets are assigned a ‘class’ based on most votes given by CF users. We have checked whether there were any instances of inconsistencies in class assignment. Furthermore, we also have checked to see if there exists a situation where equal votes are given to each class. After looping through each row and checking the data accuracy, we did not see any evidence of contradiction.

\subsection{Data Preprocessing}
The preprocessing of the text data is an essential step as it makes the raw text ready for mining, i.e., it becomes easier to extract information from the text and apply machine learning algorithms to it. If we skip this step, then there is a higher chance that you are working with noisy and inconsistent data. The main objective of this step is to clean noise of what is less relevant in order to find the sentiment of tweets such as punctuation, special characters, numbers, and terms which do not carry much weight in the text.

Data extracted from Twitter is usually not in single format and clean; due to the character limit for posts, users abbreviate words in different forms, e.g. ‘u’ instead of ‘you’. They also tend to be informal and use non-standard words (weird language), ignore grammar, avoid punctuation, or use multiple punctuation marks like ‘!!!!!’ and emojis to express feelings. Twitter is used globally, so users mix languages, they write different scripts, and write their language in roman script. Users also come up with new words and terms depending on the latest trends. Twitter is also used for ads which creates noise in data. These need to be filtered out to get true representation of the text. 

To clean the text, we took the following steps to overcome the challenges and characteristics of the data describe above. Below are a couple examples we address during preprocessing:
    \begin{description}
    \item [$\bullet$] Conversion of text to all lower-case characters
    \item [$\bullet$] Removal of special characters, “@user” etc. symbols in tweet 
    \item [$\bullet$] Conversion of different or non-standard language into English 
    \item [$\bullet$] Handling Hashtags in the tweet rather than simply removing them since Hashtags contains important information 
    \item [$\bullet$] Elimination of markups and Removal of URLs using regex 
    \item [$\bullet$] Conversion of non-textual elements like emojis into text using GLOVE. For an instance, smile face emojis, we convert that into English “Smile”
    \end{description}

We did not use some of the traditional data cleaning techniques. For example, Stemming, as it cuts off the end part of the word and can change the meaning of the word, resulting in inaccurate sentiment analysis. Therefore, we ignored such techniques and only used the ones that were necessary and suitable for our use case.

\subsection{Data Augmentation}

Data augmentation is a data oversampling technique used to increase the size of the data by adding new samples that have a similar distribution to the original data or marginally altering the original data. The data needs to be altered in a way that preserves the class label for better performance at the classification task. Since we have very imbalanced data, we want to try to augment the minority classes like “Hate” and “Offensive” data to handle the class imbalance issue in our dataset.

Data augmentation is commonly used in computer vision. In vision, you can almost certainly flip, rotate, or mirror an image without risk of changing the original label. However, in natural language processing (NLP), it is quite different. In the natural language processing (NLP) field, it is difficult to augment text due to high complexity of language. Also, not every word has a synonym. Even changing a single word, can result in completely different context. In the following paragraph will explore the main techniques that we have used to augment our training data for class “Hate” and “Offensive”.

\textbf{Synonym Word Replacement based on pre-trained Word Embedding}: We try Word Embedding based Replacement Bert model and Synonym Augmenter “Wordnet” as our main Synonym Augmenter. In basic terms, these two models will try to find a synonym for words in the tweet and replace the original with the synonym. The selection of synonyms is based on pre-trained embedding. Below is an example using Wordnet.

\begin{center}
\includegraphics[width=0.45\textwidth]{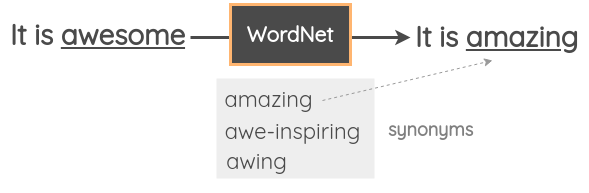}
\captionof{figure}{Synonym Word Replacement}
\end{center}

\textbf{Random Insertion}:Identifying and extracting synonyms for some randomly chosen words that are not Stop Words in the sentence. Inserting this identified synonym at some random position in the sentence. 

\begin{center}
\includegraphics[width=0.5\textwidth]{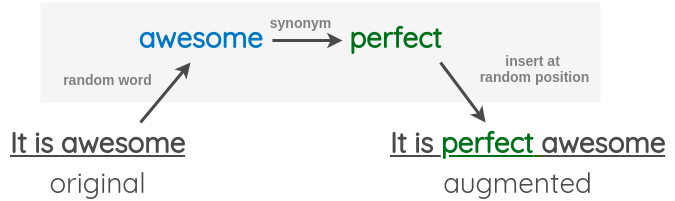}
\captionof{figure}{Random Insertion}
\end{center}

\textbf{Random Swapping}:Randomly choosing two words in the sentence and swapping their positions. This may not be a good idea in a morphologically rich language like Hindi, Marathi as it may entirely change the meaning of the sentence.

\textbf{Random Deletion}:Randomly removing word in the sentence.

For our training dataset, we try to combine some of these methods instead of using only one method. Our observation is that if we only use one method like Random Deletion, the Augmented tweet will look very similar to the original tweet. Combining these methods together gives slightly more noise to the training data but still have similar meanings. The example below shows the tweet before and after the augmentation.

\textbf{Original Tweet:}
\begin{description}
\item [$\bullet$] !!!!!!! RT @UrKindOfBrand Dawg!!!! RT @80sbaby4life: You ever fuck a bitch and she start to cry? You be confused as shit
\end{description}

\textbf{Augmented Tweet:}
\begin{description}
\item [$\bullet$] !! !! !! ! rt @ hundred urkindofbrand. .. dawg! !! ! @ fifty 80sbaby4life: ever fuck fat thought would start to cry? or you be confused as holy
\end{description}

There are many other augmentation methods like back translation, sentence shuffling, etc. We can try more different augmentation once we prove that augmentation does improve the model performance. 

Our data augmentation process applied on the training data only. At the beginning, we will split the tweet data into training and testing. After that, we will augment our minority class which is Hate and Offensive to handle the class imbalance issue. In using these methods, we would have a more balance data for training the model. 

\section{Model}

In the last few years, deep learning has had a great impact on many fields of Artificial Intelligence such as Natural Language Processing (NLP) and it has shown phenomenal improvement over standard machine learning tasks such as machine translation, text extraction, named entity relationships, Speech recognition etc. primarily because of the availability of large training data. Even though classical machine learning models are much more interpretable than deep neural networks, its main drawback is feature engineering which requires the features to be engineered manually. Shallow machine learning models such as decision trees, Naïve Bayes, KNN, random forest etc. use bag of words, n-rams, lexical features, meta data to extract the features whereas deep machine learning models use word embeddings that are fed into neural networks such as CNN, RNN, LSTM, transformers etc. The deep learning models are capable of learning the features from the raw data which helps to better extract the context for the problem on hand and hence results in much better performance. In this paper, we have focused on state-of-the art deep learning models such as LSTM and transformers. We developed our own architecture from scratch as a base case using Recurrent Neural Network and then compared its results against transfer learning from three pre-trained transformer-based models

\subsection{Albert Model} 
We tried LSTM and Bi-Directional LSTM architecture for the hate speech classification task. LSTMs have feedback connections which make them different from more traditional feedforward neural networks. This property enables LSTMs to process entire sequences of data (e.g, time series) without treating each point in the sequence independently, but rather, retaining useful information about previous data in the sequence to help with the processing of new data points. As a result, LSTMs are particularly good at processing sequences of data such as text, speech and general time-series. 

Bidirectional recurrent neural networks (RNN) are just putting two independent RNNs together. This structure allows the networks to have both backward and forward information about the sequence at every time step. Using bidirectional will run your inputs in two ways, one from past to future and one from future to past. What make this approach unique from unidirectional is that in the LSTM that runs backward you preserve information from the future and using the two combined hidden states you are able in any point in time to preserve information from both past and future. In the final model, we used Bi-Directional LSTM as BI-LSTMs show superior results since they can better understand the context.

“Hate speech dataset” was imported from a public dataset. Since the dataset was quite imbalanced for class “Hate”, we used augmentation techniques to create more examples of Hate class. After preprocessing the “tweet” column, we extracted tweets that contains the tweet bode message and “class” columns that have three target values. The dataset was split into train, validation, and test datasets. We used Keras Tokenizer to vectorize the text and convert it into a sequence of integers after restricting the tokenizer to use only topmost common 2000 words. Post padding (padding applied to the end of the sequences) was used to make the word vectors of the same size. An embedding matrix of vocabulary size and 100 dimensions was created. The embedding layer encodes the input sequence into a sequence of dense vectors of dimension embed-dim. 

\begin{center}
\includegraphics[width=0.5\textwidth]{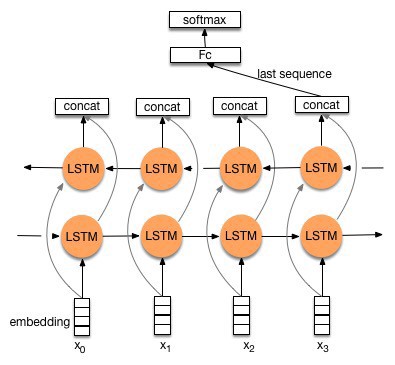}
\captionof{figure}{BI-LSTM}
\end{center}

Finally, the architecture of the Bidirectional LSTM is defined. We set the sequence length to max-length which is 512 tokens. Then, the input shape is defined. For this exercise, we tried both our own embeddings as well as pre-trained GloVe embeddings. For our embedding matrix, we initialised the embedding matrix using “uniform” initializer. Once the input shape and embedding matrix is defined, we added two layers of Bi-Directional LSTM. We used 100 memory cells in the LSTM hidden layer. The Bidirectional wrapper doubled this, creating a second layer parallel to the first, with 200 memory cells. The merge method of the Bidirectional wrapper layer was defined to concatenate both LSTM layers. Therefore a vector of 100 output values from each of the forward and backward LSTM hidden layers creates a 200-element vector output. The output of these layers is flattened. A Layer normalization layer is added to provide a uniform scale for numerical values. In this technique, normalization is applied on the neuron for a single instance across all features. Two hidden layers with ReLu activations are added to the model. The first layer has 128 neurons, and second layer has 64 neurons. To avoid over fitting, dropout regularization technique is used on both layers. We used dropout rate of 20\%.  L2 regularization penalty is applied on the hidden layer kernels. These penalties are summed into the loss function that the network optimizes. Finally, softmax layer is added to classify the tweets.  

The total number of trainable parameters on this model is 4,534,735. Then, we fit the model on the training set and check the accuracy on the validation set. We used categorical cross entropy loss function and Adam optimizer function. The batch size was set to 128 and number of epochs was set to 20 to keep the training time short. The following model loss curve was plotted. 

\begin{center}
\includegraphics[width=0.3\textwidth]{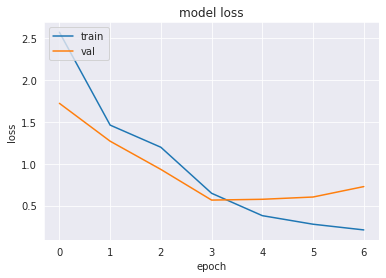}
\captionof{figure}{BI-LSTM Model Loss}
\end{center}

\subsection{Transfer Learning}
Transfer learning is a technique where a deep learning model trained on a large dataset is used to perform similar tasks on another dataset. The deep learning model in this scenario is called “pre-trained model”. It is better to use a pre-trained model as a starting point to solve a problem rather than building a model from scratch.

Transformer-based models do not need labeled data to pre-train these models. It means that we have to just provide a substantial amount of unlabeled text data to train a transformer-based model. We can use this trained model for other NLP tasks like text classification, named entity recognition, text generation, etc. This is how transfer learning works in NLP.

Transformers-based models are big neural network architectures, with a large number of parameters that can range from 100 million to over 300 million. So, training a transformer-based model from scratch on a small dataset would result in over-fitting.  This is why it is better to use a pre-trained BERT model that was trained on a huge dataset, as a starting point. We can then further train the model on our relatively smaller dataset, through a process is known as model fine-tuning.

Different Fine-Tuning Techniques:

\begin{description}
    \item [$\bullet$ Train the entire architecture]– We can further train the entire pre-trained model on our dataset and feed the output to a softmax layer. In this case, the error is back-propagated through the entire architecture and the pre-trained weights of the model are updated based on the new dataset.
    \item [$\bullet$ Train some layers while freezing others] Another way to use a pre-trained model is to train it partially. What we can do is keep the weights of initial layers of the model frozen while we retrain only the higher layers. We can try and test as to how many layers should be be frozen and how many should to be trained.
    \item [$\bullet$ Freeze the entire architecture] We can even freeze all of the layers of the model and attach a few neural network layers of our own and train this new model. It is important to note that the weights of only the attached layers will be updated during model training.
    \end{description}

In hate speech classification problem, we used the third approach. We froze all the layers of the pre-trained model during fine-tuning and append a dense layer and a softmax layer to the architecture. “Hate speech dataset” was imported from a public dataset and augmented to address class imbalance for class “Hate”. After preprocessing the “tweet” column, we extracted “tweet” that contains the tweet bode message and “class” columns that has three target values. The dataset was split into train, validation, and test datasets.

We tried three different pre-trained models for this exercise by importing Bert Base (110 million parameters), DistilBert (66 million parameters) and GPT-2 (1.5 billion parameters)

\begin{center}
\includegraphics[width=0.20\textwidth]{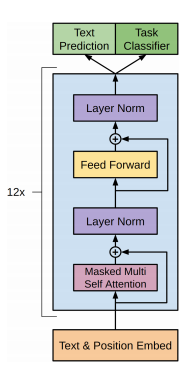}
\captionof{figure}{GPT-2 Architecture }
\end{center}

We tokenized the sentences to prepare the inputs for the model. For DistillBert and Bert, we used Hugging Face auto tokenizer and for GPT-2, we used Hugging Face GPT-2 tokenizer. For consistency purposes, the architecture of all three models was kept same. All of these tokenizers output a dictionary of two items ‘input-ids’ and “attention-mask”. We used padding to ensure that all of the messages have the same length through maximum sequence length. Truncation was set to “True” to limit the tokens to the maximum sequence length.

Then, the architecture of the model is defined using Tensor flow Keras package. Sequence length was set to “Max-Length” for all three models. All the layers of DistilBert, Bert and GPT-2 pre-trained model were frozen. The shapes of inputs and attention masks were defined. GloVE pre-trained embedding with 100 dimensions were used. After shaping the inputs and masks, we set the output of GPT-2 pre-trained model to the last hidden layer. We then flattened the output of last hidden layer of the pre-trained model. A Layer normalization layer is added to provide a uniform scale for numerical values. In this technique, normalization is applied on the neuron for a single instance across all features. Two trainable hidden layers were added with RELu activation function. The first layer has 128 neurons, and second layer has 64 neurons. To avoid over fitting, dropout regularization technique is used on both layers. We used a dropout rate of 20\%.  L2 regularization penalty is applied on the hidden layer kernels. These penalties are summed into the loss function that the network optimizes. Finally, we added softmax layer for the model to classify the output. “Categorical Cross Entropy” loss function was used. We chose “Adam” optimizer function for the model. 

Then, we fit the model on training set and check the accuracy on validation set. We used categorical cross entropy loss function and Adam optimizer function. The batch size was set to 128 and number of epochs was set to 20 to keep the training time short. The following model loss curves and evaluation metrics were plotted for all three transfer learning models.

\begin{figure}
\centering
  \centering
  \includegraphics[width=0.45\linewidth]{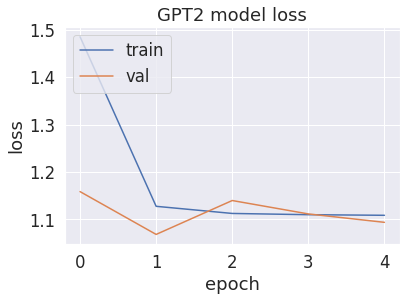}
  \label{fig:sub1}
  \centering
  \includegraphics[width=0.45\linewidth]{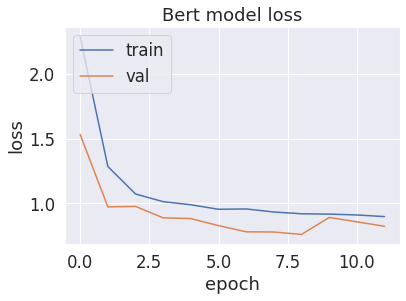}
  \label{fig:1}
  \centering
  \includegraphics[width=0.45\linewidth]{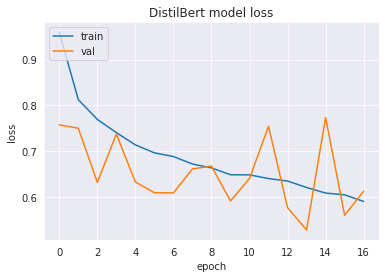}
  \label{fig:1}
\caption{Transfer Learning Model Loss}
\label{fig:test}
\end{figure}

In term of the accuracy on testing data, we found that the Base Model (Albert Model) shows the best performance compared to all three transfer learning models. We can see the detail model comparison in Figure 8.

\begin{center}
\includegraphics[width=0.4\textwidth]{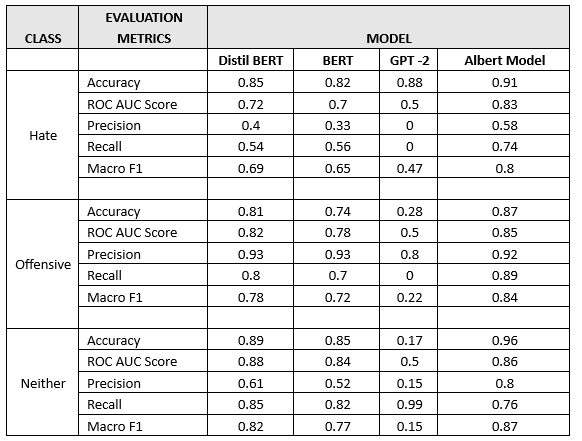}
\captionof{figure}{Model Comparison }
\end{center}

\subsection{Further Hyper-Parameters Tuning}
We experimented with tuning different hyper-parameters on the base model (Albert Model) by using Keras tuner library. The following hyper parameters were tuned in the find model:
\begin{description}
    \item [$\bullet$]We set a minimum value of 32 and a maximum value of 512 for memory cells in both LSTM hidden layers with step value of 32.
    \item [$\bullet$]Minimum number of neurons in both dense hidden layers was set to 32 and the maximum number was set to 512 with step of 32 neurons.
    \item [$\bullet$]We tried dropout rate of 0.2, 0.35, 0.5, 0.65 and 0.8 for both Bi Directional LSTM layers as well as two dense layers.
    \item [$\bullet$]Finally, we tried three learning rates of 0.01, 0.001 and 0.0001.
    \end{description}

The following optimal setting were obtained from the tuned model:
\begin{description}
    \item [$\bullet$]352 Hidden cells and dropout rate of 0.65 for the first LSTM layer
    \item [$\bullet$]320 Hidden cells and dropout rate of 0.80 for the second LSTM layer
    \item [$\bullet$]288 neurons and dropout rate of 0.2 for the first dense layer
    \item [$\bullet$]416 neurons and dropout rate of 0.35 for the second dense layer
    \item [$\bullet$]The optimal learning rate for the optimizer was 0.0001
    \end{description}

\begin{center}
\includegraphics[width=0.4\textwidth]{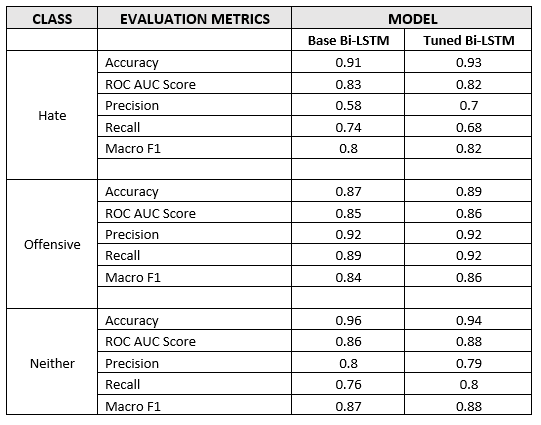}
\captionof{figure}{Model Comparison After Hyper-Parameters Tuning}
\end{center}

%%%%%%%%%%%%%%%%%%%%%%%%%%%%%%%%%%%%%%%%%%%%%%%%%%%%%%%%%%%%%%%%%%%%%%
\section{Results Interpretation}

In the application of this model, we realize that some prediction errors will be more costly to a brand than others. The two major errors we are focused on for the business use case of this model include falsely classifying regular content as hate speech and falsely omitting hate speech and thus classifying it as regular content. We focused on continuing to improve model training to minimize errors, though it is important to understand the economic implications of these prediction errors to help optimize the model. We use a cost benefit matrix and confusion matrix below to help quantify and assess the model errors. 

False positive is defined as a tweet that was classified as hate speech by the model, but was in fact a normal, acceptable tweet. False negative is defined as a tweet that was classified as acceptable by the model, but was in fact content that included some form of hate speech. 

True negative is defined as a tweet that was classified as not hate speech or offensive language, and was in fact a normal, acceptable tweet. Our model is not built to maximize good tweets. While general tweets obviously reflect economic value for a brand, for the purpose of our analysis we define the true negative as zero.

True positive is defined as a tweet that was classified as hate speech and is in fact hate speech content. The business use case for our model is prevention of lost revenue. In this case, a true positive is associated with detecting potential future losses (the tweet can be flagged and deleted before any damage is done). We will explore how to quantify the benefit of hate speech detection in the next section, but for the purpose of this analysis we now define it as zero.

\subsection{GitHub Repository}
We use Github to track and manage our code development process. This is the link to the public GitHub repository:
https://github.com/BenchengW/Offensive-Language-Detection-DL

In our GitHub documentation, there is a notebook called “Instructions.ipynb”. This notebook contains all of the necessary steps to show how to user Albert Model from Scratch to detect a text as Hate, Offensive or Neither. Here is the link to the notebook:
https://github.com/BenchengW/Offensive-Language-Detection-DL/blob/main/documentation/Instructions.ipynb

\subsection{Acknowledgement }
The work was divided evenly amongst all the team members. Sub-teams were created to research, design and develop different components for this paper. Atsu and Hafiza were responsible for the preprocessing component and preparing the documentation. Similarly, Hafiza and Jason designed, trained and tested the main model built using TensorFlow, and prepared the documentation. Ajay and Jason completed the transfer learning portion and prepared the documentation. Natalie prepared the report for executive summary, business value and model production. Ben designed the tweet sentiment analysis, data augmentation, prepare the documentation and kept tracking the code version on Github. After each of the components were completed by the sub-teams, Ben compiled the individual write ups into a report. 

%%%%%%%%%%%%%%%%%%%%%%%%%%%%%%%%%%%%%%%%%%%%%%%%%%%%%%%%%%%%%%%%%%%%%%

\end{document}